\pdfoutput=1

\documentclass[11pt]{article}

\usepackage[final,preprint]{acl}
\usepackage{times}
\usepackage{latexsym}

\usepackage[T1]{fontenc}

\usepackage[utf8]{inputenc}

\usepackage{microtype}

\usepackage{inconsolata}

\usepackage{graphicx}
\usepackage{listings}
\usepackage{booktabs}
\usepackage{color}
\usepackage{float}

\definecolor{dkgreen}{rgb}{0,0.6,0}
\definecolor{gray}{rgb}{0.5,0.5,0.5}
\definecolor{mauve}{rgb}{0.58,0,0.82}

\title{ParsiPy: NLP Toolkit for Historical Persian Texts in Python}

\author{
 \textbf{Farhan Farsi\textsuperscript{1}}, 
 \textbf{Parnian Fazel\textsuperscript{2}}, 
 \textbf{Sepand Haghighi\textsuperscript{3}}, 
 \textbf{Sadra Sabouri\textsuperscript{3,4}}, \\
 \textbf{Farzaneh Goshtasb\textsuperscript{5}}, 
 \textbf{Nadia Hajipour\textsuperscript{5}}, 
 \textbf{Ehsaneddin Asgari\textsuperscript{6}}, 
 \textbf{Hossein Sameti\textsuperscript{7}} \\ 
 \textsuperscript{1}Amirkabir University of Technology, 
 \textsuperscript{2}University of Tehran, 
 \textsuperscript{3}Open Science Laboratory, \\
 \textsuperscript{4}University of Southern California,
 \textsuperscript{5}Institute for Humanities and Cultural Studies, \\
 \textsuperscript{6}Qatar Computing Research Institute, 
 \textsuperscript{7}Sharif University of Technology \\ 
 \small{\texttt{farhan1379@aut.ac.ir, parnian.fazel@ut.ac.ir, sepand@openscilab.com, sabourih@usc.edu,}}\\ \small{\texttt{f.goshtasb@ihcs.ac.ir, n.hajipour@ihcs.ac.ir, easgari@hbku.edu.qa, sameti@sharif.edu}}
}

\begin{document}

\maketitle
\begin{abstract}
    The study of historical languages presents unique challenges due to their complex orthographic systems, fragmentary textual evidence, and the absence of standardized digital representations of text in those languages.
    Tackling these challenges needs special NLP digital tools to handle phonetic transcriptions and analyze ancient texts.
    This work introduces ParsiPy\footnote{\url{https://github.com/openscilab/parsipy}}, an NLP toolkit designed to facilitate the analysis of historical Persian languages by offering modules for tokenization, lemmatization, part-of-speech tagging, phoneme-to-transliteration conversion, and word embedding.
    We demonstrate the utility of our toolkit through the processing of Parsig (Middle Persian) texts, highlighting its potential for expanding computational methods in the study of historical languages. Through this work, we contribute to computational philology, offering tools that can be adapted for the broader study of ancient texts and their digital preservation.
\end{abstract}
\section{Introduction}
Ancient languages serve as windows into the past, offering valuable insights into human history and the evolution of communication. The connection between language and culture has long been recognized, with scholars using ancient languages such as Greek~\cite{ostwald2011language}, Italian~\cite{lomas2013language}, and Latin~\cite{farrell2001latin} to uncover the social contexts of historical civilizations. These languages not only preserve cultural heritage but also provide a lens for studying the development of linguistic structures and thought patterns~\cite{kaplan2013cultural}. Despite significant advancements in Natural Language Processing (NLP), which have transformed the study of modern languages, the application of these technologies to ancient languages remains underexplored~\cite{magueresse2020low}. As preliminary attempts, some researchers have tailored NLP tools developed for Pre-modern English~\cite{johnson2021classical} and Sumerian~\cite{guzman2023introducing}, yet many historically significant languages, such as Old Persian and Middle Persian (P\=ars\=ig), still lack sufficient computational resources and tools. Expanding NLP research to include these underserved languages can help bridge critical gaps in historical linguistics while contributing to the preservation of invaluable cultural knowledge.

P\=ars\=ig, represents one such language~\cite{haug1870essay}.
Despite its historical importance as a bridge between ancient Iranian languages and modern Persian (see Appendix \ref{sec:app:parsig-lang} for more details), P\=ars\=ig has received minimal attention in computational linguistics.
Its challenges include a highly limited digital corpus, complex writing system variations, and the absence of standardized computational resources for processing P\=ars\=ig texts in the originally written form.

To address this gap, we introduce ParsiPy, the first NLP toolkit in Python specifically designed for processing P\=ars\=ig.
Our framework includes tools for word embeddings, lemmatization, tokenization, and part-of-speech (POS) tagging. Our POS tagging system includes three models—Hidden Markov Model with Viterbi decoding, logistic regression, and random forest.

P\=ars\=ig was written in multiple scripts, but the Book Pahlavi script, widely used in Zoroastrian texts, lacks a Standard Unicode encoding\footnote{\url{https://www.unicode.org/standard/unsupported.html}}.
As a result, most digital resources rely on phonetic transcriptions.
To address this, ParsiPy includes a phoneme-to-transliteration module with rule-based and LSTM models.
We also provide a tool for converting this transliteration to Book Pahlavi.
Future work could develop Unicode support, enabling broader computational applications.

ParsiPy addresses the challenges of processing P\=ars\=ig texts by using rule-based and statistical methods, which are more effective than large language models for this low-resource language. As a foundational NLP toolkit, ParsiPy enhances computational analysis of P\=ars\=ig, supports digital research, and serves as a model for similar efforts in other ancient languages. The code is available as a Python package on GitHub.

In this paper, we outline the structure of the P\=ars\=ig language in Section \ref{sec:parisg-lan}, followed by related works on NLP toolkits for ancient languages in Section \ref{sec:rw}. Section \ref{sec:method} details the system design of ParsiPy, while Section \ref{sec:dataset} describes the dataset used for training and evaluation. We then assess our toolkit in Section \ref{sec:eval} and discuss future research directions in Section \ref{sec:discuss}.

\begin{figure}
    \centering
    \includegraphics[width=\linewidth]{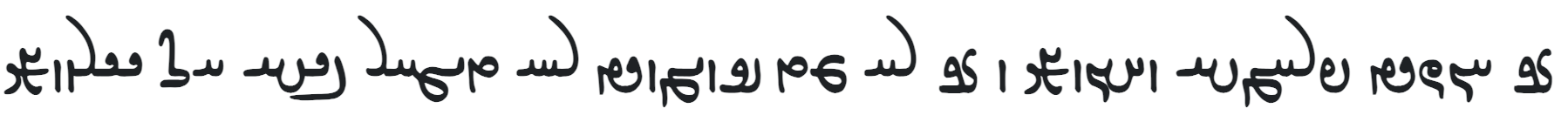}
    \caption{An example of P\=ars\=ig text in original written form from \textit{Andarze Azarabade Mehrsepandan} database. It reads ``\textit{ān uzīd frāmōš kun ud ān nē mad ēstēd rāy tēmār bēš ma bar}'' and it means ``\textit{Forget what is gone and do not worry about what has not yet come.}''}
    \label{fig:parsipy-exampletext}
\end{figure}

\section{P\=ars\=ig Language Structure}
\label{sec:parisg-lan}
P\=ars\=ig, the language of the Sassanian Empire (224–651 CE), an ancestor of modern Persian (Farsi), has a unique linguistic structure that can be divided by specific features in phonology, morphology, syntax, and orthography. In this section, we go through its specific characteristics.

\textbf{Phonology and Orthography}
of P\=ars\=ig is similar to that of modern Persian, though there are important historical phonetic differences.

The P\=ars\=ig alphabet consists of only fourteen letters to represent the entire range of sounds, as illustrated in the Appendix. Consequently, several letters possess multiple phonetic values. This variation in phonetic values presents challenges in reading P\=ars\=ig.The difficulty is further compounded by the different shapes the letters can take, depending on their position in the word \cite{MacKenzie1971}.
A significant portion of P\=ars\=ig words is written using Aramaeograms (known as uzwārišn), where words of Aramaic origin are spelled using P\=ars\=ig characters \cite{Goshtasb2023}.

\textbf{Morphology}
of P\=ars\=ig is primarily inflectional, with both nouns and verbs marked for grammatical roles such as case, tense, and mood.
P\=ars\=ig originally had two cases: one reserved for the grammatical subject, and the other for all other syntactic functions (oblique). 
These cases are commonly referred to as the `direct' case (used for the subject of the sentence) and the `oblique' case (used for objects, indirect objects, and other syntactic functions)~\cite{Skjaervoe2005}.
Verbs are inflected for various grammatical features such as tense, mood, and person~\cite{Brunner1977}.
Additionally, P\=ars\=ig verbs often include the use of verbal particles and suffixes to convey different meanings and functions, which can make morphological analysis complex.
Another common feature is enclitic pronouns, short pronoun-like elements that attach to words to show possession or objects, which can make segmentation tasks complex.

\textbf{Syntax}
of P\=ars\=ig follows a Subject-Object-Verb (SOV) word order \cite{Dabir-Moghaddam2014}, but this structure is flexible depending on context or emphasis.
This variability makes syntactic parsing more challenging. The language also uses prepositions and postpositions, and relative clauses often form with subordinators, requiring tools to detect clause boundaries accurately.

\textbf{Semantic and Lexical Features. }
The vocabulary of P\=ars\=ig includes many loanwords from Aramaic~\cite{4aefba25-6aea-3776-808a-4bc419c55714}, which creates challenges for distinguishing between native and borrowed words.
Also, due to the script’s lack of vowel markings, polysemy (words with multiple meanings) and tomography (identical spellings with different pronunciations) present challenges.
These features complicate tasks like word sense disambiguation and machine translation.

Developing NLP tools for P\=ars\=ig requires addressing these unique linguistic features. Techniques such as character-level models for handling logograms, graph-based parsing for non-fixed word order, and morphological analyzers for suffix-rich structures can be particularly effective.
This paper uses an excerpt from a Zoroastrian manuscript~\cite{Goshtasb2022Justice}, originally written in P=ars=ig, as an example. Figure \ref{fig:parsipy-exampletext} shows the original handwritten text. The passage is from \textit{Andarze Azarabade Mehrsepandan}, a collection of life advice, with an English translation: \textit{Forget what is gone and do not worry about what has not yet come.} This example was chosen for its variety of words, characters, and POS tags. The phonetic transcription is as follows:
\begin{lstlisting}
s='ān uzīd frāmōš kun ud ān nē mad ēstēd rāy tēmār bēš ma bar'
\end{lstlisting}
\section{Related Work}
\label{sec:rw}
\textbf{NLP on Ancient Languages.}
Despite the growing interest in computational approaches for ancient languages~\cite{vico2023larth,long2023development}, the P\=ars\=ig language remains entirely unexplored in this domain.
Farsi itself is classified as a low-resource language~\cite{shamsfard2019challenges}, and ancient Farsi, such as P\=ars\=ig, suffers from even greater limitations.
To the best of our knowledge, except for ~\cite{rahnamoun-rahnamoun-2025-semantic} who recently presented a set of word embeddings for P\=ars\=ig language, work on this language is scarce. 
These limitations include the lack of annotated corpora, standardized scripts, and linguistic resources.
Existing efforts in the broader area of ancient language processing have focused on better-documented languages.
For instance, \cite{sahala2023neural} developed a neural pipeline for POS-tagging and lemmatization of Cuneiform languages, while \cite{vico2023larth} introduced resources for Etruscan machine translation.
Similarly, ~\cite{naaijer2023transformer} proposed a transformer-based parser for Syriac morphology, demonstrating the applicability of modern NLP techniques to ancient scripts.

\textbf{Tools for Ancient Languages.}
In the broader domain of tool development for ancient languages, \cite{guzman2023introducing} introduced an open-source library for Sumerian text analysis, while \cite{koch-etal-2023-tailored} presented a handwritten text recognition system for Medieval Latin manuscripts.
Recognizing the unique challenges of ancient languages, researchers like \cite{johnson-etal-2021-classical} have developed toolkits to simplify their processing and bridge initial research gaps. These open-source toolkits are especially valuable, streamlining foundational tasks and making further research more accessible.
Another prominent example is DadmaTools, a comprehensive open-source NLP toolkit for Modern Farsi that supports tokenization, lemmatization, and part-of-speech tagging~\cite{jafari-etal-2025-dadmatools}.
However, ancient languages like P\=ars\=ig require additional considerations, such as handling non-standardized scripts, logograms, and transcription-transliteration tasks. 

These works highlight the challenges and opportunities in processing ancient languages while emphasizing the importance of creating specialized tools for their unique linguistic and orthographic features.
Addressing the lack of research on Middle Persian, ParsiPy is the first computational framework for the language, featuring transcription-transliteration module and morphological analyzers to tackle its challenges.

\begin{figure*}
    \centering
    \includegraphics[width=\linewidth]{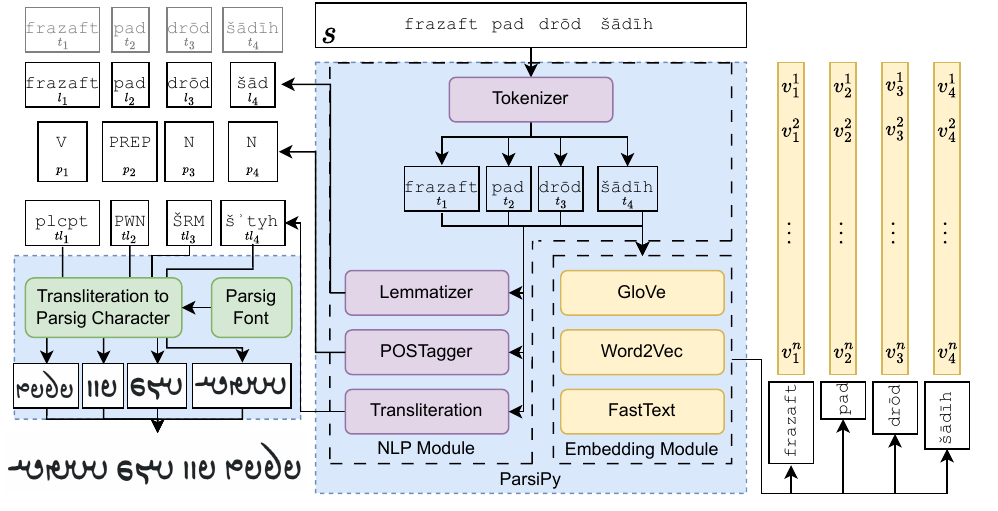}
    \caption{Parsipy Framework Overview. Input string $s$ goes into tokenized into $n$ tokens ($t_1, \cdots,t_n$) and the embedding module would generate word embeddings for each token ($v_1, \cdots, v_n$). Lemmatizer extracts the lemma for each token ($l_1,\cdots,l_n$), POSTagger tags each token with its part-of-speech in the sentence ($p_1, \cdots,p_n$), and Transliteration module (Phoneme to Transliteration: P2T) generates a middle-form representation of tokens by which they can transform into Parsig in hand-written form. The example sentence is from Corpus Of Pahlavi Texts (Jamaspji Dastur Minochehrji Jamasp Asana) which is gathered and translated by \cite{Goshtasb2022Justice}. The English translation of it is ``It ended with greetings (= happiness)'' and it is chosen for the sake of simplicity.}
    \label{fig:parsipy-overview}
\end{figure*}

\section{System Design}
\label{sec:method}
The ParsiPy toolkit is built upon three main components: the embedding module, the NLP task modules, and Parsig character generator.
The first component provides a semantical representation for words and sentences while the second one analyzes the input sentence syntactically.
Since there is no well-known Unicode representation for the Parsig language we decided to set the input to Parsipy modules as a more well-accepted form of this language which is phonetics representation.
However, we present a middle form (transliteration) which can be used to be converted into Parsig characters.

An overview of the ParsiPy structure is presented in Figure \ref{fig:parsipy-overview}.
The blue dotted parts are current works' contributions. Yellow boxes are embedding modules, purple boxes are NLP modules and green boxes are Parsig character generator modules.
Tokenized input can be passed to the embedding module to get embeddings for each token, Lemmatizer, POS Tagger and Transliteration yield lemmization, part-of-speech tagger, and transliteration of each token.
Transliterations can be converted to chunks of originally Parsig character set and hence stack together to form sentences in Parsig original form.

\subsection{Embedding Module}
We integrated support for state-of-the-art embedding methods for textual data, including FastText~\cite{bojanowski2017enriching}, GloVe~\cite{pennington2014glove}, and Word2Vec~\cite{church2017word2vec}, which are well-suited for minimal data sizes, aligning with prior works on low-resource tasks~\cite{nazir2022toward,gaikwad2020adaptive,saadatinia2025enhancing,fesseha2021text}.
Parsipy's embedding module enables the transformation of words and sentences into continuous vector spaces.
These vector representations capture semantic relationships between words, facilitating downstream tasks such as word similarity~\cite{islam2008semantic}, sentiment analysis~\cite{medhat2014sentiment}, and text classification~\cite{kowsari2019text}.
They also enable models to identify patterns, improving performance on tasks like document clustering~\cite{article}, and question answering.

\subsection{NLP modules}
The embedding module focuses on the semantic aspects of language, while other ParsiPy components handle syntactical analysis through tasks like part-of-speech tagging, offering insights into grammatical relationships. We included key NLP tasks—tokenization, lemmatization, and part-of-speech tagging—with easy-to-use APIs for researchers. Additionally, we provide a middle-form transliteration of Parsig, which can be converted into its original character representation. 

We developed a pipeline API that covers NLP tasks, including phoneme representation to transliteration, with a usage and output style similar to the Stanza toolkit~\cite{qi2020stanza}.

\begin{lstlisting}
from parsipy import pipeline, Task
result = pipeline(sentence=s, tasks=[Task.TOKENIZER, Task.LEMMA, Task.POS, Task.P2T])
\end{lstlisting}

We now explain each part separately showcasing Parsipy's output to give better insights on the matter.
The output of the above code fills \verb|result| with a dictionary with a field for each of the provided tasks, i.e., \verb|Task.TOKENIZER| (tokenization), \verb|Task.LEMMA| (lemmatization), \verb|Task.POS| (part-of-speech tagging), and \verb|Task.P2T| (transcription to transliteration).

\textbf{Tokenizer.}
Tokenization is the process of transforming sentences into smaller units, such as words or sub-words like word pieces and byte pairs~\cite{mielke2021between}.
Effective tokenization is particularly important for historical and low-resource languages like Parsig, where complex morphology and script variations present unique challenges.

For the tokenization module in ParsiPy, we developed a SentencePiece unigram language model~\cite{kudo2018subword} with a vocabulary size of 40,000 tokens.
We chose SentencePiece because it operates directly on raw text without requiring predefined word boundaries, making it particularly suitable for Parsig with inconsistent or non-standardized orthography.
Additionally, its subword-based approach helps efficiently handle out-of-vocabulary words and rare morphological variations which ensures better adaptability for low-resource languages with limited digital resources.

The tokenized version of our example sentence is shown below.
To enhance identification, we assign a unique token ID to each token, making it easier for traceability during analysis.
\begin{lstlisting}
[
    {'id': 0, 'text': 'ān'},
    {'id': 1, 'text': 'uzīd'},
    {'id': 2, 'text': 'frāmōš'},
    {'id': 3, 'text': 'kun'},
    {'id': 4, 'text': 'ud'},
    {'id': 5, 'text': 'ān'},
    {'id': 6, 'text': 'nē'},
    {'id': 7, 'text': 'mad'},
    {'id': 8, 'text': 'ēstēd'},
    {'id': 9, 'text': 'rāy'},
    {'id': 10, 'text': 'tēmār'},
    {'id': 11, 'text': 'bēš'},
    {'id': 12, 'text': 'ma'},
    {'id': 13, 'text': 'bar'}
]
\end{lstlisting}

\textbf{Lemmatizer.}
Lemmatization reduces words to their canonical form, using linguistic rules and context, unlike stemming, which simply removes affixes~\cite{khyani2021interpretation}.
This is particularly essential for historical languages like Parsig, where inflectional variations and complex morphology require a more nuanced approach to text normalization.
In ParsiPy, considering the static nature of the Parsig language and its fixed vocabulary size, we constructed a comprehensive table to store the lemma of each word. Additionally, we formulated linguistic rules to effectively handle specific cases, particularly compound words. This approach facilitated the development of a rule-based lemmatization module that accurately determines the lemma for each word in a text by applying linguistic rules tailored to the Parsig language.
Our approach accounts for common morphological transformations.
In the following example, the lemma of \textit{ēstēd} is extracted as \textit{ēst}, while other words remain unchanged.
For out-of-vocabulary words, the original word itself is returned as the lemma.

\begin{lstlisting}
[
    {'lemma': 'ān', 'text': 'ān'},
    {'lemma': 'uzīd', 'text': 'uzīd'},
    {'lemma': 'frāmōš', 'text': 'frāmōš'},
    {'lemma': 'kun', 'text': 'kun'},
    {'lemma': 'ud', 'text': 'ud'},
    {'lemma': 'ān', 'text': 'ān'},
    {'lemma': 'nē', 'text': 'nē'},
    {'lemma': 'mad', 'text': 'mad'},
    {'lemma': 'ēst', 'text': 'ēstēd'},
    {'lemma': 'rāy', 'text': 'rāy'},
    {'lemma': 'tēmār', 'text': 'tēmār'},
    {'lemma': 'bēš', 'text': 'bēš'},
    {'lemma': 'ma', 'text': 'ma'},
    {'lemma': 'bar', 'text': 'bar'}
]
\end{lstlisting}

\textbf{Part of Speech Tagger.}
Part-of-speech (POS) tagging is the task of assigning grammatical roles to words in a sentence.
POS tags aid downstream tasks such as syntactic parsing, machine translation, and information retrieval.
We have support for three different POS taggers: I) HMM \& Viterbi model, II) Logistic Regression model, and III) Random Forest Classifier model. We evaluated them on our dataset and reported the results in Section \ref{sec:eval}.

Our POS tagger supports a complete tag set for Parsig, covering categories such as nouns (\verb|N|), adjectives (\verb|ADJ|), verbs (\verb|V|), adverbs (\verb|ADV|), pronouns (\verb|PRO|), prepositions (\verb|PREP|), postpositions (\verb|POST|), conjunctions (\verb|CONJ|), determiners (\verb|DET|), numerals (\verb|NUM|), particles (\verb|PART|).
Additionally, we incorporated morphological features unique to Parsig, such as automatic recognition of adverbial suffixes (e.g., \textit{īhā}) and verb conjugation patterns.
We report the performance of the POS tagger on these different categories in Section \ref{sec:eval}.

The following example illustrates the output generated by our POS tagging module. For instance, the word \textit{uzīd}, which means \textit{go}, should be tagged as a verb. 

\begin{lstlisting}
[
    {'POS': 'DET', 'text': 'ān'},
    {'POS': 'V', 'text': 'uzīd'},
    {'POS': 'N', 'text': 'frāmōš'},
    {'POS': 'V', 'text': 'kun'},
    {'POS': 'CONJ', 'text': 'ud'},
    {'POS': 'DET', 'text': 'ān'},
    {'POS': 'ADV', 'text': 'nē'},
    {'POS': 'V', 'text': 'mad'},
    {'POS': 'V', 'text': 'ēstēd'},
    {'POS': 'POST', 'text': 'rāy'},
    {'POS': 'N', 'text': 'tēmār'},
    {'POS': 'N', 'text': 'bēš'},
    {'POS': 'ADV', 'text': 'ma'},
    {'POS': 'N', 'text': 'bar'}
]
\end{lstlisting}

\textbf{Phoneme to Transliteration (P2T) Module.} 
Parsig is predominantly represented in a phonemic script in digital resources.
Transliteration is a representation in middle form between the phonetic representation and the actual Parsig character set.
Therefore, transliterations are crucial components of Parsig linguistic processing since they bridge between these two modalities.
A key challenge in this domain is bridging the gap between phonemic representation and standardized transliteration.
In our approach, we explored rule-based models, as data scarcity limits the effectiveness of data-driven machine-learning methods.
By leveraging linguistic rules specific to Parsig, we developed a robust system that produces high-quality transliteration.  

We used character sets from \textit{Huzwāreš}, borrowed from Aramaic~\cite{goshtasb2021corpus}, as initial transliterations, represented in capital letters in ParsiPy's output (e.g., \texttt{ZK} for \texttt{ān}). This exploration refines our model, enhancing accuracy in Parsig text representation.

\begin{lstlisting}
[
    {'translite': 'ZK', 'text': 'ān'},
    {'translite': 'ʾwcyt', 'text': 'uzīd'},
    {'translite': 'plʾmwš', 'text': 'frāmōš'},
    {'translite': 'OḆYDWNty', 'text': 'kun'},
    {'translite': 'W', 'text': 'ud'},
    {'translite': 'ZK', 'text': 'ān'},
    {'translite': 'LA', 'text': 'nē'},
    {'translite': 'mt', 'text': 'mad'},
    {'translite': "YKOYMWyt'", 'text': 'ēstēd'},
    {'translite': 'lʾd', 'text': 'rāy'},
    {'translite': 'tymʾl', 'text': 'tēmār'},
    {'translite': 'byš', 'text': 'bēš'},
    {'translite': 'AL', 'text': 'ma'},
    {'translite': 'YḆLWN', 'text': 'bar'}
]
\end{lstlisting}

\subsection{Transliteration to Written Form}
To encourage the use of the Parsig language in the original form we present an enhanced version of the Parsig font and an executable file for converting translation into a written format of Parsig language texts with the original character set using that font.

\textbf{Parsig Font.}
We have refined and expanded an existing font set for the Parsig alphabet. This involved adjusting the positioning of letters relative to the baseline to achieve better alignment. The enhanced version of the font set is included in the supplementary materials.

\textbf{Transliteration to Parsig Character Module.}
Additionally, we introduce an executable tool that converts Parsing sentences from their transliterated form, aligned with ParsiPy's output formats, into their original script using this font set.

\section{Dataset (Parsig Database)}
\label{sec:dataset}
\begin{table}[ht]
    \centering
    \begin{tabular}{l r}
        \toprule
        \textbf{Statistic} & \textbf{Value} \\
        \midrule
        Total Documents & 120 \\
        Total Words & 93,518 \\
        Unique Tokens & 8,839 \\
        Distinct Lemmas & 4,641 \\
        \bottomrule
    \end{tabular}
    \caption{Summary of the P\=ars\=ig Database}
    \label{tab:Parsig_summary}
\end{table}
We used P\=ars\=ig Database as a comprehensive resource for P\=ars\=ig texts, meticulously curated by domain experts (two authors from this work) with advanced linguistic backgrounds for our training and evaluation.
It contains 120 documents with a total of 93,518 words, including 8,839 unique tokens and 4,641 distinct lemmas (Table \ref{tab:Parsig_summary}).
Each entry is carefully annotated with multiple linguistic layers, such as lemmatization, part-of-speech (POS) tagging, and transliterations. 
The data set also includes translations in both English and Persian for its usability for researchers studying the evolution of the Ps\=ars\=ig language and its relationship with modern Persian.

The project was initiated in December 2018 and officially launched in 2020 with an initial corpus of around 40,000 words.
Over time, the database has expanded, and it remains an ongoing initiative aimed at further enriching Parsig linguistic resources. 
The P\=ars\=ig Database adheres to strict annotation standards, including transcriptions, transliteration preserving original spellings, and \textit{Huzvāreš} annotations for ideographic forms. It is accessible for research in Persian language processing\footnote{\url{https://www.Parsigdatabase.com/}}.

\section{Evaluation}
\label{sec:eval}
To ensure the quality of the models used in PasriPy, we evaluated our models using texts from the P = ars = ig database, which includes content from well-known books in that language.
Our dataset for evaluation consists of texts from the following books: ~\citet{jamaspasana1913pahlavi}, ~\citet{dhabhar1930andarj}, ~\citet{anklesaria1913danak}, and~\citet{anklesaria1935kar}.

\textbf{Metrics.}
ParsiPy consists of modules for different tasks that require different metrics for evaluation.
For the P2T module, due to its resemblance to the P2G (phoneme-to-grapheme) module, we measured performance using Word Error Rate (WER)~\cite{klakow2002testing} and Character Error Rate (CER)~\cite{morris2004} which are type of Levenshtein distance~\cite{levenshtein1966binary} in word and character level respectively.
Comparing two strings (one predicted and one actual) then projected down to finding the number of substitutions $S$, deletions $D$, and insertions $I$ needed to change one to another and the error rate is calculated as follows, where $N$ is the total number of parts (words or characters) in both two strings.

\begin{equation}
    ER = \frac{S + D + I}{N}
\end{equation}

For the Lemmatizer module, accuracy is used to assess performance, reflecting the proportion of words correctly lemmatized into their base forms out of the total words evaluated.

For the POS tagger, we used standard accuracy, precision, recall, and F1-score as our evaluation metrics since POS tagging is fundamentally a token classification task.
Here, we report the evaluation results of various parts of the ParsiPy framework across different models.

\subsection{P2T}
ParsiPy's rule-based P2T module yielded 29.764\% WER and 13.525\% CER on the P\=ars\=ig dataset.
We also tested other methods for P2T which we
report the results in Table \ref{tab:p2t-eval} (The lower WER and CER, the better it is).

\begin{table}[h]
    \centering
    \begin{tabular}{l c c}
        \toprule
        \textbf{Model}  & \textbf{WER} & \textbf{CER} \\ \midrule
        Rule-based model& \textbf{29.764} & \textbf{13.525} \\
        LSTM & 31.009 & 22.125 \\
        \bottomrule
    \end{tabular}
    \caption{P2T Models Performance on the P\=ars\=ig Dataset}

    \label{tab:p2t-eval}
\end{table}

\subsection{Lemmatizer} The Lemmatizer module of the ParsiPy toolkit achieved an accuracy of 0.894, indicating that 89.4\% of the words were correctly reduced to their base forms during the evaluation.

\subsection{POS Tagger}

Given the limited dataset for POS tagging as a multi-class classification task, we initially handcrafted linguistic features, a common approach in data-scarce settings~\cite{lee-lee-2023-lftk,shumilov2024data}.
We then experimented with foundational machine learning models such as logistic regression and random forest, following methodologies used by other researchers working with small datasets~\cite{jahara2020towards,ashrafi2024optimizing,liao2007logistic}.
We split the dataset into training and test sets, using 10\% of the data for testing.
We also tried other methods for POS Tagging, which are presented in Table \ref{tab:pos-eval}.
Finally, we compared these models' performance with our heuristic approach, which uses an HMM-based model and a Viterbi decoder for POS tag prediction.
\begin{table}[htbp]
    \centering
    \resizebox{\linewidth}{!}{
    \begin{tabular}{l c c c c}
    \toprule
    \textbf{Model} & \textbf{Accuracy} & \textbf{F1} & \textbf{Recall} & \textbf{Precision$^*$} \\
    \midrule
    Viterbi & 0.98319 & 0.74465 & 0.70933 & 0.89071 \\
    Logistic Regression & \textbf{0.98984} & 0.8213 & 0.81977 & \textbf{0.93396} \\
    Random Forest Classifier & 0.98874 & \textbf{0.84832} & \textbf{0.9268} & 0.9268 \\
    \bottomrule
    \end{tabular}
    }
    \caption{The performance comparison of the different POS tagger models is presented, with all metrics reported as macro averages, except for Precision, which is reported in micro due to the absence of some classes, rendering the macro Precision score unavailable.}
    \label{tab:pos-eval}
\end{table}
\begin{table*}[ht]
    \centering
    \resizebox{\textwidth}{!}{
    \begin{tabular}{lccccccccccccc}
        \toprule
            & \textbf{ADJ} & \textbf{ADV} & \textbf{CONJ} & \textbf{DET} & \textbf{EZ} & \textbf{N} & \textbf{NUM} & \textbf{PART} & \textbf{POST} & \textbf{PREP} & \textbf{PRON} & \textbf{Unknown} & \textbf{V} \\
        \midrule
        \textbf{ACC} & 0.97273 & 0.98359 & 0.98518 & 0.99325 & 0.99431 & 0.96056 & 0.99735 & 0.99947 & 0.99682 & 0.99457 & 0.98968 & 0.99907 & 0.98703 \\
        \textbf{AUC} & 0.86275 & 0.91874 & 0.96979 & 0.9433 & 0.98586 & 0.96373 & 0.95177 & 0.9373 & 0.86387 & 0.98768 & 0.89477 & 0.61111 & 0.97294 \\
        \textbf{F1} & 0.80675 & 0.87321 & 0.95345 & 0.87531 & 0.95459 & 0.93786 & 0.94118 & 0.77778 & 0.78182 & 0.97468 & 0.83884 & 0.36364 & 0.94912 \\
        \textbf{Precision} & 0.8977 & 0.90466 & 0.95983 & 0.86058 & 0.93388 & 0.90612 & 0.9816 & 0.7 & 0.84314 & 0.97048 & 0.89035 & 1.0 & 0.94421 \\
        \textbf{Recall} & 0.73254 & 0.84387 & 0.94715 & 0.89055 & 0.97624 & 0.97191 & 0.90395 & 0.875 & 0.72881 & 0.97891 & 0.79297 & 0.22222 & 0.95407 \\
        \bottomrule
    \end{tabular}}
    \caption{Performance metrics for different POS classes with Random forest POS Tagger and Random Forest Classifier. Accuracy (ACC) Macro = 0.98984, F1 Macro = 0.84832. The evaluation was conducted using the PyCM library~\cite{haghighi2018pycm}.}
    \label{tab:pos-metrics-rf}
\end{table*}

\textbf{Features.}
As hand-crafted features for the input of the POS tagger, we incorporated the following attributes of each word: the string representation of the word itself, whether it ends with \textit{\=ih\=a} (indicating adverb), whether it is the first or last word of the sentence, the string representation of the previous and next words, the first two and last two characters of the word, the first and last character as prefixes and suffixes, the tag of the previous word in the sentence, and the word length.

\textbf{Models.}
We implemented three models for the POS tagger classification model.
First, we implemented a Hidden Markov Model~\cite{eddy1996hidden} with the Viterbi decoding algorithm~\cite{forney1973viterbi} for sequence labeling.
This model relies on emission probabilities (word-to-tag likelihoods) and transition probabilities between adjacent tags.
To handle out-of-vocabulary words, we applied Laplace smoothing with a constant of $0.001$.
For the other two models we fed the feature representations of the sentences into a \verb|DictVectorizer| pipeline~\cite{pedregosa2011scikit} to obtain vector representations, which were subsequently used to train our baseline POS taggers with two foundational machine learning classifiers: \verb|LogisticRegression| and \verb|RandomForestClassifier|.

To optimize performance, we conducted a grid search over a wide range of hyperparameters, evaluating models using 10-fold cross-validation. The best hyperparameters for the logistic regression model were \verb|penalty='l2'|, \verb|C=1.0|, and \verb|solver='lbfgs'|, while for the random forest model, they were \verb|n_estimators=100|, \verb|criterion='gini'|, \verb|min_samples_split=2|, and \verb|min_samples_leaf=1|.
These two models outperformed our baseline hubristic model and the random forest POS tagger yielded a slightly higher f1-score (0.84832). Class-based performance of this classifer has been presented in Table \ref{tab:pos-metrics-rf}.

While some of the categories like Numbers are easy to detect for our model due to their nature, other categories like particles were harder to detect due to the low presentation rate in the training data.
For a more detailed analysis of various model evaluations, please refer to Appendix \ref{sec:app:pos_class_results}.



\section{Discussion}
\label{sec:discuss}
We now outline potential future directions for advancing NLP research in low-resource languages and particularly P\=ars\=ig.

\textbf{Expandability of ParsiPy.}
Due to its modular design, ParsiPy offers a flexible framework that allows researchers to integrate new tasks and train additional models, improving the accuracy of existing functionalities.
The exploratory path we followed in developing this library can serve as a foundational scaffolding for other researchers aiming to build an NLP toolkit for low-resource languages.
To facilitate this process and ensure easier integration, we will open-source the training code and toolkit package. This approach enables researchers to seamlessly build upon our work, and with transparent ML model transportation frameworks like Pymilo~\cite{rostami2024pymilo}, these models can be deployed and served effectively.
Community engagement and collaboration will be key to refining and expanding ParsiPy's capabilities.

\textbf{Parsig Unicode.}
One of the next steps in enhancing resources for the P\=ars\=ig language is establishing a standardized Unicode representation.
To our knowledge, previous attempts at Unicode representation have remained incomplete or faced significant challenges, and currently, there is no standard Unicode for this script.
A future direction is to develop a Unicode standard for P\=ars\=ig that accounts for both intra-language character similarities and cross-language relations, improving encoding quality and enhancing P\=ars\=ig's accessibility for linguistic research.

Furthermore, the creation of linguistic resources, such as annotated corpora and lexicons, will significantly enhance computational efforts for this historically significant language. By providing structured datasets, we aim to facilitate NLP advancements, ensuring better text processing, character recognition, and model training for Parsig.

\section{Limitations}  
Our work has certain limitations. While we concentrated on fundamental NLP tasks to establish a strong foundation for the Parsig language, some tasks, such as Named Entity Recognition (NER), were not included in this phase of development. Expanding support for these tasks remains an important goal for future iterations of our work.

Additionally, the scarcity of high-quality annotated data posed a significant challenge. Due to these limitations, we were unable to fully leverage state-of-the-art transformer-based models, which have demonstrated superior performance over traditional approaches in various NLP applications. Addressing this data gap would allow us to explore more advanced architectures.

Despite these constraints, we are committed to the continued development of ParsiPy. In future work, we plan to expand its capabilities to support a broader range of NLP tasks, incorporate cutting-edge deep learning techniques, and perform a more comprehensive error analysis. By refining our methodologies and leveraging new data sources, we aim to improve the accuracy, robustness, and overall effectiveness of ParsiPy for the research community and practical applications.

\section{Conclusion}
ParsiPy provides a vital NLP toolkit for analyzing P\=ars\=ig texts, addressing challenges like the lack of computational tools.
With modules for tokenization, lemmatization, part-of-speech tagging, and phoneme-to-grapheme conversion, it facilitates linguistic analysis and digital preservation. By combining rule-based and statistical methods, ParsiPy proves effective for low-resource languages and serves as a model for similar efforts. Future work could enhance transliteration accuracy, expand deep learning models, and develop Unicode support for Book Pahlavi, further advancing historical linguistics and computational philology.


\bibliography{custom}
\appendix
\section{Parsig Language}
\label{sec:app:parsig-lang}
\subsection{Overview of Middle Iranian Languages}
The Middle Iranian languages span a long period (about 1,200 years) from the fall of the Achaemenid Empire to the 9th century CE.
Written documents from this period exist in six languages: Middle Persian (Sasanian Pahlavi or P\=ars\=ig), Parthian Pahlavi, Sogdian, Bactrian, Khotanese, and Khwarezmian.
Among these, P\=ars\=ig is particularly significant, as it is the precursor to modern Persian and the only Iranian language with written records from its ancient phase, including Old Persian inscriptions.

Parsig was the language of Zoroastrian Middle Persian texts, Sasanian inscriptions, Manichaean writings, and Christian Middle Persian texts, while each written in different scripts.
The majority of surviving P\=ars\=ig texts are religious Zoroastrian writings, composed in the Book Pahlavi script, also known as cursive Pahlavi.

\textbf{Zoroastrian Middle Persian Texts.}
The surviving Pahlavi texts encompass a wide range of topics, with the largest category being Zand texts—translations and interpretations of the Avesta into Pahlavi—along with works derived from these interpretations, such as \textit{Dēnkard}, \textit{Bundahišn}, \textit{Selections of Zādspram}, \textit{Dādestān ī Dēnīg}, and \textit{Pahlavi Rivayats}.

Beyond these, Pahlavi literature includes various other genres: philosophical-theological works like \textit{Škand Gumānīg Wizār} and \textit{Pas Dānišn-kāmag}; mystical and prophetic texts such as \textit{Ardā Vīrāz-nāmag} and \textit{Jāmāsp’s Prophecies}; ethical and didactic literature including \textit{Yādgar ī Buzurgmihr} and \textit{Dādestān ī Mēnōg ī Xrad}; debates and boastful compositions like \textit{The Assyrian Tree}; historical and geographical accounts such as \textit{Kārnāmag ī Ardaxšīr ī Pāpakān} and \textit{Šahrestān-hā ī Ērān}; epic literature like \textit{Yādgar ī Zarērān}; and legal texts including \textit{Šāyest nē Šāyest} and \textit{Mādayān ī Hazār Dādestān}.
Additionally, educational treatises, such as \textit{Xusraw ud Rēdag} and \textit{The Chess and Nard Treatise}, and lexicons like the \textit{Pahlavi Lexicon} further enrich the corpus.

These texts are invaluable for understanding Iran’s cultural, religious, and historical heritage, while their linguistic analysis significantly contributes to Persian language studies, historical linguistics, and lexical research \cite{durkin2004, macuch2009, tafazzoli1999, amouzgar1994}

\begin{figure}[h]
    \centering
    \includegraphics[width=1\linewidth]{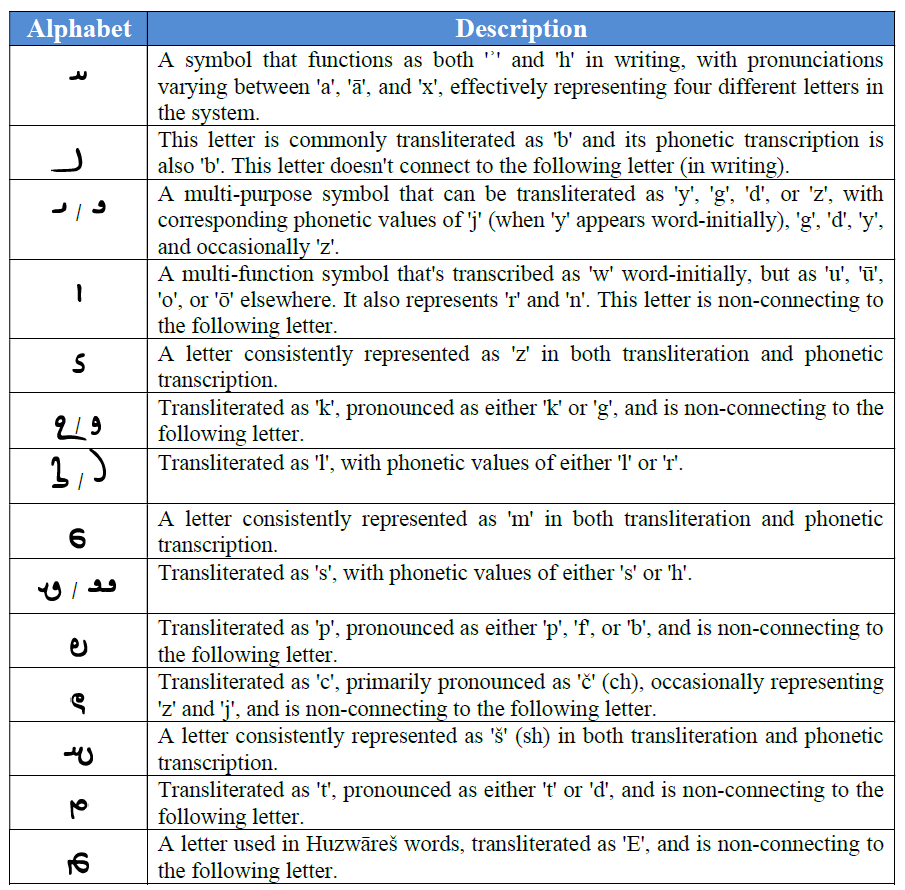}
    \caption{The 14 letters of the P\=ars\=ig alphabet, used in the Middle Persian language.}
    \label{fig:parsig-letters}
\end{figure}

\subsection{The Book Pahlavi Script}
All Western Middle Iranian scripts originate from the Aramaic script and were used to write Parthian and Middle Persian (Sasanian Pahlavi) texts. The major script variations include Parthian (Inscriptional Pahlavi), used for Parthian inscriptions and early Sasanian texts; Inscriptional Pahlavi, which appeared in royal and noble inscriptions of the Sasanian period; Book Pahlavi, primarily used for Zoroastrian Middle Persian texts; and Psalter Pahlavi, employed in Middle Persian Christian texts. The script specifically used for Zoroastrian Middle Persian writings is called \textit{Book Pahlavi}, a cursive script referred to by Islamic-era writers as \textit{ram dabīra} or \textit{hām dabīra}, meaning ``common script.'' \textit{Book Pahlavi} consists of 14 letters and is written from right to left (shown in Figure \ref{fig:parsig-letters}).

\begin{table*}[ht]
    \centering
    \resizebox{\textwidth}{!}{
    \begin{tabular}{lccccccccccccc}
        \toprule
            & \textbf{ADJ} & \textbf{ADV} & \textbf{CONJ} & \textbf{DET} & \textbf{EZ} & \textbf{N} & \textbf{NUM} & \textbf{PART} & \textbf{POST} & \textbf{PREP} & \textbf{PRON} & \textbf{Unknown} & \textbf{V} \\
        \midrule
        \textbf{ACC} & 0.97168 & 0.97929 & 0.95391 & 0.98878 & 0.98691 & 0.95003 & 0.99532 & 0.99893 & 0.99693 & 0.99492 & 0.98424 & 0.99947 & 0.98103 \\
        \textbf{AUC} & 0.8396 & 0.8774 & 0.95 & 0.92651 & 0.95571 & 0.95006 & 0.9092 & 0.5 & 0.81452 & 0.99032 & 0.7865 & 0.6 & 0.94332 \\
        \textbf{F1} & 0.78842 & 0.81258 & 0.86293 & 0.7931 & 0.90909 & 0.92438 & 0.89489 & 0.0 & 0.77228 & 0.9757 & 0.69271 & 0.33333 & 0.92102 \\
        \textbf{Precision} & 0.9316 & 0.87047 & 0.79444 & 0.73516 & 0.89908 & 0.9 & 0.98675 & None & 1.0 & 0.96705 & 0.86928 & 1.0 & 0.95063 \\
        \textbf{Recall} & 0.68339 & 0.7619 & 0.94435 & 0.86096 & 0.91932 & 0.95012 & 0.81868 & 0.0 & 0.62903 & 0.98452 & 0.57576 & 0.2 & 0.8932 \\
        \bottomrule
    \end{tabular}}
    \caption{Performance metrics for different POS classes with HMM \& Viterbi POS Tagger. Accuracy (ACC) Macro = 0.98319, F1 Macro = 0.74465}
    \label{tab:pos-metrics-viterbi}
\end{table*}

\begin{figure}
    \centering
    \includegraphics[width=\linewidth]{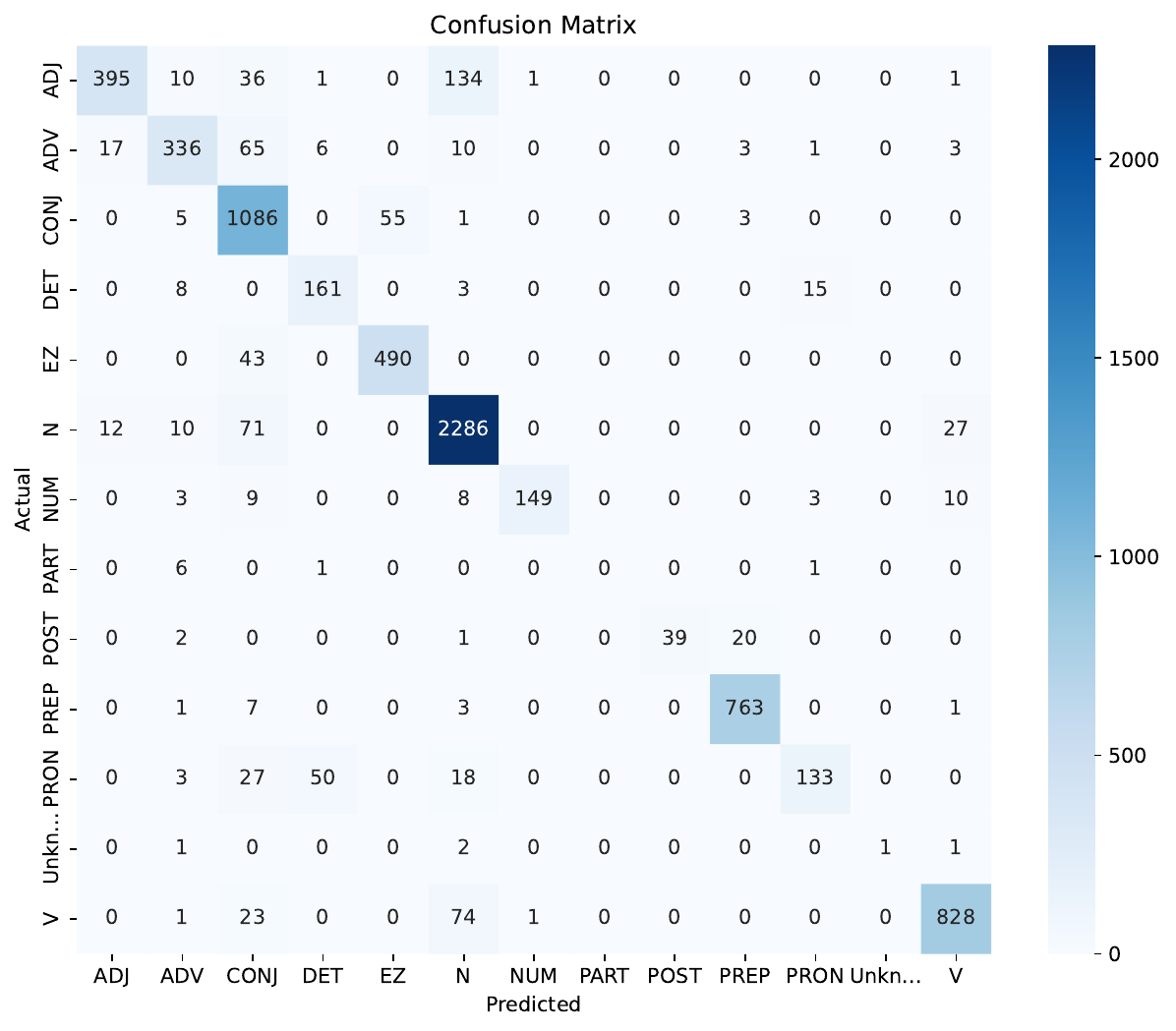}
    \caption{Confusion matrix for HMM \& Viterbi}
    \label{fig:app_cm_vm}
\end{figure}

\begin{table*}[ht]
    \centering
    \resizebox{\textwidth}{!}{
    \begin{tabular}{lccccccccccccc}
        \toprule
            & \textbf{ADJ} & \textbf{ADV} & \textbf{CONJ} & \textbf{DET} & \textbf{EZ} & \textbf{N} & \textbf{NUM} & \textbf{PART} & \textbf{POST} & \textbf{PREP} & \textbf{PRON} & \textbf{Unknown} & \textbf{V} \\
        \midrule
        \textbf{ACC} & 0.97361 & 0.98568 & 0.98583 & 0.99533 & 0.99367 & 0.96396 & 0.99925 & 0.99864 & 0.99623 & 0.99397 & 0.99291 & 0.9991 & 0.98975 \\
        \textbf{AUC} & 0.89234 & 0.91405 & 0.97065 & 0.97119 & 0.99119 & 0.96251 & 0.98873 & 0.92255 & 0.80157 & 0.99179 & 0.90472 & 0.5 & 0.97665 \\
        \textbf{F1}  & 0.82893 & 0.87214 & 0.95347 & 0.91014 & 0.95281 & 0.94589 & 0.98141 & 0.70968 & 0.73684 & 0.97294 & 0.85449 & 0.0 & 0.95813 \\
        \textbf{Precision} & 0.86531 & 0.91525 & 0.95821 & 0.87709 & 0.91974 & 0.93384 & 0.98507 & 0.61111 & 0.94595 & 0.95739 & 0.90196 & None & 0.95695 \\
        \textbf{Recall} & 0.7955 & 0.8329 & 0.94877 & 0.94578 & 0.98834 & 0.95826 & 0.97778 & 0.84615 & 0.60345 & 0.989 & 0.81176 & 0.0 & 0.95931 \\
        \bottomrule
    \end{tabular}}
    \caption{Performance metrics for different POS classes with Logistic regression POS Tagger and Logistic Regression. Accuracy (ACC) Macro = 0.98984, F1 Macro = 0.8213}
    \label{tab:pos-metrics-lr}
\end{table*}
\begin{figure}
    \centering
    \includegraphics[width=\linewidth]{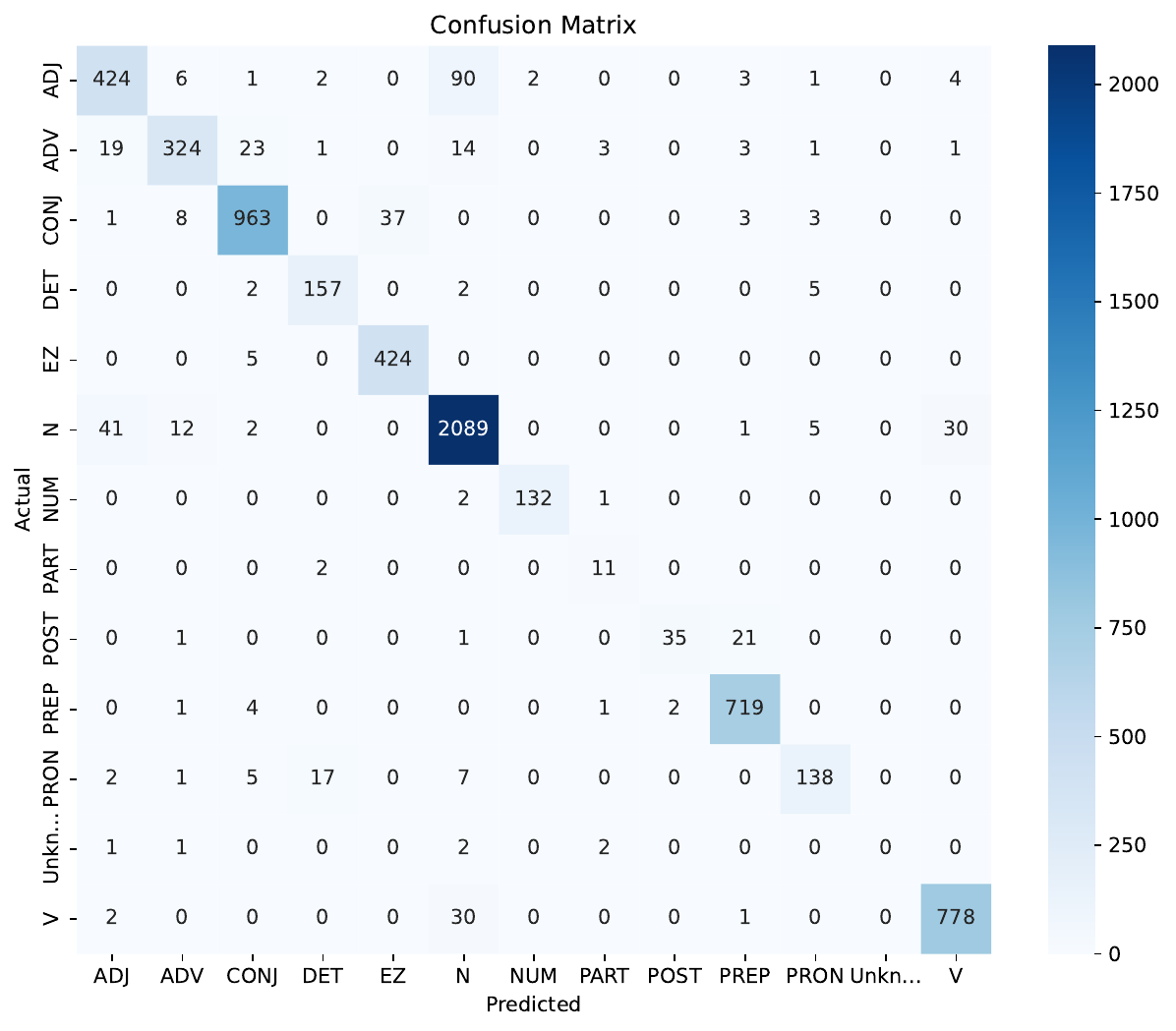}
    \caption{Confusion matrix for Logistic regression}
    \label{fig:app_cm_lr}
\end{figure}

\begin{figure}
    \centering
    \includegraphics[width=\linewidth]{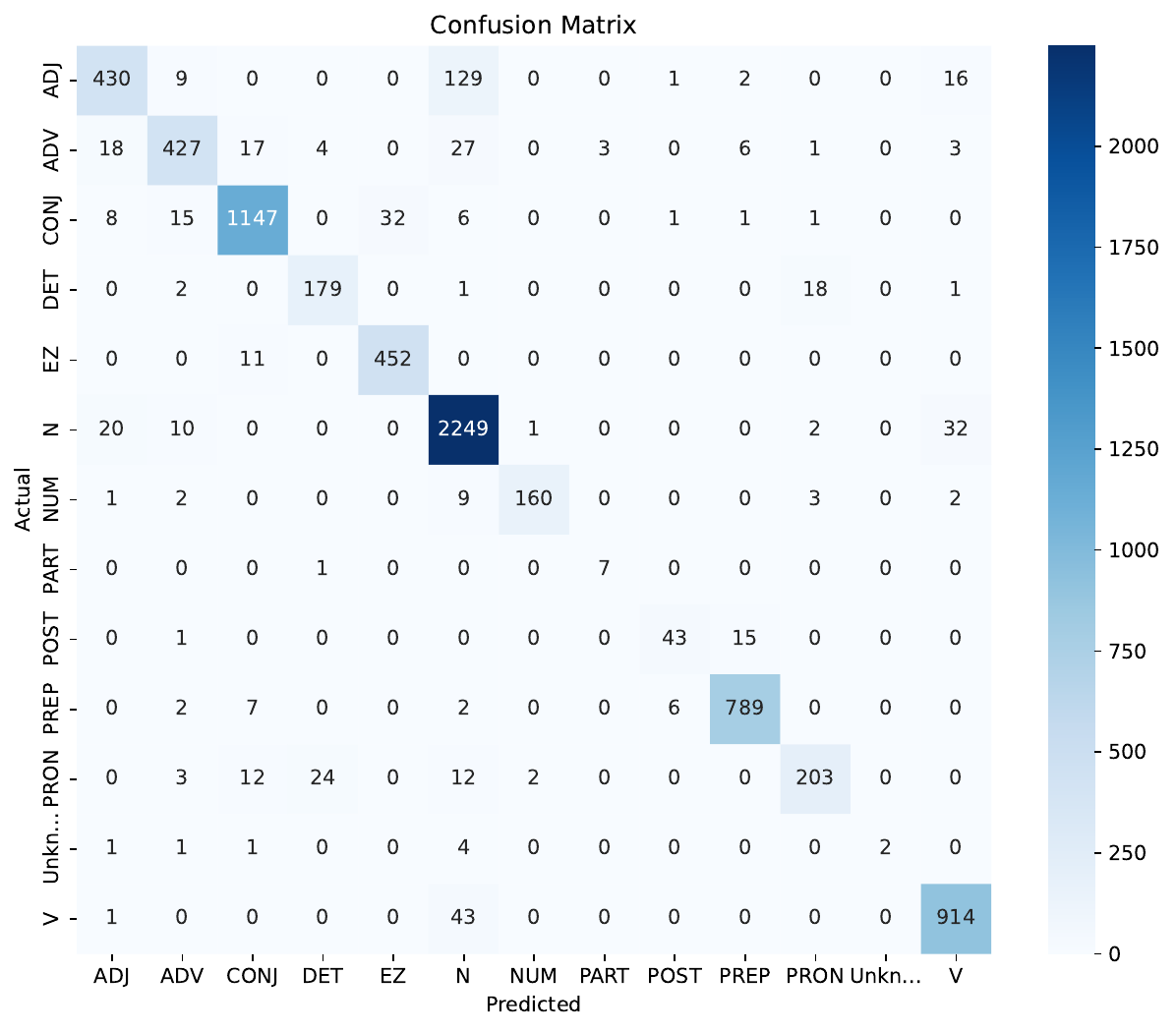}
    \caption{Confusion matrix for Random forest}
    \label{fig:app_cm_rfc}
\end{figure}





\section{POS Tagger Classification Results}
In this part we present class-based metrics confusion matrices for POS tagger classifiers.
The evaluation was conducted using the PyCM library~\cite{haghighi2018pycm}.
\label{sec:app:pos_class_results}

\subsection{HMM \& Viterbi}
Table \ref{tab:pos-metrics-viterbi} represent class-based metrics for HMM \& Viterbi POS tagger and confusion matrix is presented in Figure \ref{fig:app_cm_vm}.

\subsection{Logistic Regression}
Table \ref{tab:pos-metrics-lr} represent class-based metrics for Logistic regression POS tagger and confusion matrix is presented in Figure \ref{fig:app_cm_lr}.

\subsection{Random Forrest Classifier}
Table \ref{tab:pos-metrics-rf} represent class-based metrics for Random forest POS tagger and confusion matrix is presented in Figure \ref{fig:app_cm_rfc}.

\end{document}